\documentclass[a4paper, 10pt, conference]{ieeeconf}      
\usepackage{FG2024}

\FGfinalcopy 

\IEEEoverridecommandlockouts                              
\usepackage{graphicx} 
\usepackage{amsmath} 
\usepackage{multirow}
\usepackage{svg}
\usepackage{subcaption}

\def\FGPaperID{268} 

\title{\LARGE \bf
Social-MAE: A Transformer-Based Multimodal Autoencoder for Face and Voice
}

\author{\parbox{16cm}{\centering
    {\large Hugo Bohy$^1$, Minh Tran$^2$, Kevin El Haddad$^1$, Thierry Dutoit$^1$ and Mohammad Soleymani$^2$}\\
    {\normalsize
    $^1$ Numediart Institute, ISIA Lab, University of Mons, Mons, Belgium\\
    $^2$ Institute for Creative Technologies, University of Southern California, Los Angeles, CA, USA}}
    \thanks{The work was partially sponsored by the Army Research Office and was accomplished under Cooperative Agreement Number W911NF-20-2-0053. The views and conclusions contained in this document are those of the authors and should not be interpreted as representing the official policies, either expressed or implied, of the Army Research Office or the U.S. Government. The U.S. Government is authorized to reproduce and distribute reprints for Government purposes notwithstanding any copyright notation herein.}
}

\usepackage{fancyhdr}
\begin{document}

\ifFGfinal
\thispagestyle{empty}
\pagestyle{empty}
\else
\author{Anonymous FG2024 submission\\ Paper ID \FGPaperID \\}
\pagestyle{plain}
\fi
\maketitle

\thispagestyle{fancy}

\begin{abstract}
Human social behaviors are inherently multimodal necessitating the development of powerful audiovisual models for their perception. 
In this paper, we present Social-MAE, our pre-trained audiovisual Masked Autoencoder based on an extended version of Contrastive Audio-Visual Masked Auto-Encoder (CAV-MAE), which is pre-trained on audiovisual social data. Specifically, we modify CAV-MAE to receive a larger number of frames as input and pre-train it on a large dataset of human social interaction (VoxCeleb2) in a self-supervised manner. We demonstrate the effectiveness of this model by fine-tuning and evaluating the model on different social and affective downstream tasks, namely, emotion recognition, laughter detection and apparent personality estimation. The model achieves state-of-the-art results on multimodal emotion recognition and laughter recognition and competitive results for apparent personality estimation, demonstrating the effectiveness of in-domain self-supervised pre-training. Code and model weight are available here https://github.com/HuBohy/SocialMAE.


\end{abstract}

\section{INTRODUCTION}
\label{INTRO}
Human emotions and social behaviors are expressed and perceived through multiple modalities. While verbal communication can provide information on a person's communicative intent and emotions, non-verbal communication has shown to be equally or even more important~\cite{knapp1978nonverbal}. Socially intelligent systems require multimodal methods allowing them to perceive human social and expressive behaviors. Understanding expressions and social behaviors can be achieved by analyzing audiovisual modalities, i.e., face, body and voice. Although unimodal approaches, e.g., vision from facial expression or audio for tracking arousal, can reach a high performance~\cite{cai2023marlin, gong2021ast}, fusing two modalities increases the efficiency and robustness of multimodal systems~\cite{nagrani2021attention, fayek2020large}. 
Information from different modalities can be congruent and reinforce each other for more effective communication. For instance, a smiling face aligned with a cheerful tone reinforce the expression of happiness. Information from different modalities can also be complementary, improving clarity and reducing uncertainty, such as a confident tone coupled with a smiling face. There is also a possibility of interaction between modalities generating new meanings from contradictory information from different modalities, e.g., expression of irony with conflicting face and voice behaviors.  

In supervised learning, the availability of labeled data is often limited by laborious annotation. A common solution is to fine-tune a pre-trained model, i.e., to use models that have previously been trained on a larger dataset of similar nature~\cite{dosovitskiy2020image, peebles2023scalable, hassaniNeighborhoodAttentionTransformer2023}. Self-supervised learning has been proposed to leverage large-scale unlabeled datasets to pre-train a model by pre-training models using a pretext task. One such pretext task is autoencoding, i.e., encoding input information into an often more compact latent representation and decoding it back to the original space. The encoder module of an autoencoder can be re-used as a pre-trained model with powerful and discriminative representations to be fine-tuned or re-used for downstream tasks. 

Past work extensively explored the natural interactions between audio and visual signals for representation learning \cite{owens2016visually, owens2016ambient, owens2018audio, ng2020you2me, afouras2020self, furnari2020rolling, shi2022learning} through self-supervision with a variety of pretext tasks. Synthesis-based strategies \cite{shi2022learning, owens2016visually, owens2016ambient} have been proposed, where audio and visual signals are artificially combined to facilitate learning cross-modal associations. Alignment-based methods \cite{korbar2018cooperative, owens2018audio, arandjelovic2017look, gao2020visualechoes}, on the other hand, focus on aligning signals from both modalities in time or space, aiming to extract meaningful correlations between them. Another line of research involves the application of masked autoencoding (MAE)~\cite{georgescu2023audiovisual}, where the model learns to reconstruct the missing portions of either the audio or visual input, fostering representation learning through learning the structure of the data.  
Recently, two models, namely MAViL~\cite{huang2024mavil} and CAV-MAE~\cite{gong2022contrastive}, have explored the combination of MAE with contrastive learning and demonstrated state-of-the-art (SOTA) performance on audio-visual classification. Adding contrastive learning allows the models to learn inter-modality representations.

\begin{figure*}[thpb]
  \centering
  \includegraphics[width=0.95\textwidth]{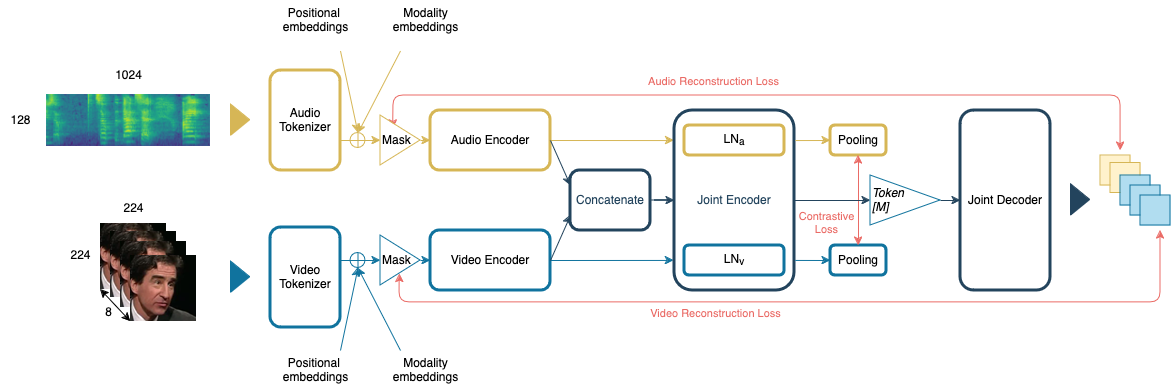}
  \caption{Social-MAE model for voice and face analysis in videos. The model is pre-trained to reconstruct audio and visual modalities from masked portions of their corresponding input, narrowing the difference between each modality representation.}
  \label{fig:Social-MAE}
  \vspace{-10pt}
\end{figure*}

Despite the popularity of emotion and social behavior perception, datasets for such tasks are often limited in size due to the high cost of labeling. Most existing audiovisual methods are based either on transfer learning with models trained on out-of-domain data, e.g., AudioSet~\cite{gemmekeAudioSetOntology2017a}, or trained from scratch. However, the desired input data should contain human faces and voices. Existing audiovisual encoders, e.g., \cite{gong2022uavm}, also lack the temporal fidelity in the visual domain. In contrast, expressive behaviors in the human face are rather dynamic and fast-moving. There is limited work, e.g., \cite{saamlTran}, on audiovisual encoders suitable for the automatic perception of human emotional and communicative behaviors. 

In this paper, we present Social-MAE, a pre-trained audiovisual model based on Masked Autoencoder. We aim to adapt a self-supervised method with superior results on audio event recognition for the audiovisual understanding of human social behaviors. We evaluate our model against several baselines on three different social and affective tasks: emotion recognition, laughter detection and apparent personality estimation. The main contributions of this work are as follows. 
\begin{itemize}
    \item We present Social-MAE, a model based on CAV-MAE architecture adapted to social context by pre-training on a large-scale social dataset;
    \item To develop Social-MAE, we modify CAV-MAE to accept multiple frames providing higher temporal fidelity at visual input;
    \item Our experiments demonstrate the importance of in-domain pre-training for affective and social tasks. Our model reaches or outperforms SOTA models on relevant tasks.
\end{itemize}


\section{METHOD}
\label{METHO}
In this section, we present Social-MAE (Fig.~\ref{fig:Social-MAE}), an adapted version of CAV-MAE that focuses on voice and face. The model is composed of two modality-specific encoders followed by a joint encoder module and a joint decoder module. Each module relies on a set of Transformer layers~\cite{vaswani2017attention} made of an attention block, a feed-forward network, residual connections and layer normalization~\cite{ba2016layer}. We describe the pre-processing pipeline in Sec.~\ref{METH:Preproc}, the model overview in Sec.~\ref{METH:CAVMAE} and the self-supervised training in Sec.~\ref{METH:SSL}.

\subsection{Audiovisual Tokenization}
\label{METH:Preproc}
The architecture follows a mid-fusion scheme: both audio and video are first encoded in two separate branches for several encoder layers before merging into a joint encoder. Audio data are pre-processed as in CAV-MAE: we convert the input audio waveform into a sequence of 128-dimensional log Mel filterbank features computed with a 25 ms Hamming window and an overlap of 10 ms. We pad or crop the length of the input to keep 1024 audio frames, resulting in a $128\times 1024$ spectrogram. The spectrogram is processed as an image that we split in N $16\times16$ non-overlapping patches. Each patch is projected with a linear layer to a 1-dimensional embedding of size 768, referred to as a \textit{token}. We add a trainable positional embedding to each token to provide information about the token order. 

Visual inputs differ from CAV-MAE as they consist of eight randomly selected frames as proposed in~\cite{huang2024mavil} rather than single frame. Each frame is an RGB image of the face bounding box scaled to $224\times224$ pixels, resulting in a $8\times224\times224\times3$ video input. We split the video into N $2\times16\times16$ patches with no overlap, flatten and project with a linear layer into tokens of size 768. A trainable positional embedding is added to each token as well. 
Another trainable parameter provides information about each token's modality and weights the modality's importance. After adding positional and modality embeddings, a random mask with a rate of p\% is applied to the input tokens, providing the model only with (1-p)\% of the original audio and/or video sequence.

\subsection{Model Description}
\label{METH:CAVMAE}
This section presents an overview of the autoencoder architecture as described in ~\cite{gong2022contrastive}. The model first processes an input sequence in separate encoders, each leveraging unimodal information. The modality encoders are stacks of 11 Transformer layers that aim to encode internal patterns in the input sequence. The joint encoder comprises a single Transformer layer on top of the modality encoders. Each modality is processed by the respective encoder followed by the joint encoder either individually or concatenated with the second modality depending on the targeted loss. The layer normalization on top of the joint encoder differs for audio, video and multimodal processing. 
The weights of the joint encoder are shared regardless of its input modality, as it was shown that weight sharing lightens the model without degrading performance~\cite{lee2020parameter}. 
The unimodal tokens are averaged following the average pooling method, while the multimodal tokens are fed to the joint decoder, which is a stack of 8 Transformer layers. It aims to retrieve the original video and audio from an input sequence made of the encoded tokens and a learnable token \textbf{M} repeated at masked positions. The reconstruction loss is the distance between tokens \textbf{M} at the output of the decoder and their corresponding original tokens.

\begin{figure*}[thpb]
  \centering
  \begin{subfigure}{.5\textwidth}
      \centering
      \includegraphics[width=.95\textwidth]{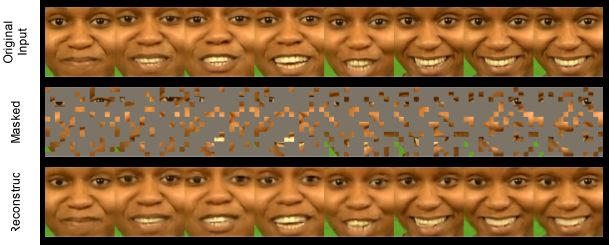}
      \caption{Reconstruction on CREMA-D.}
      \label{fig:sfig1}
    \end{subfigure}%
    \begin{subfigure}{.5\textwidth}
      \centering
      \includegraphics[width=.95\textwidth]{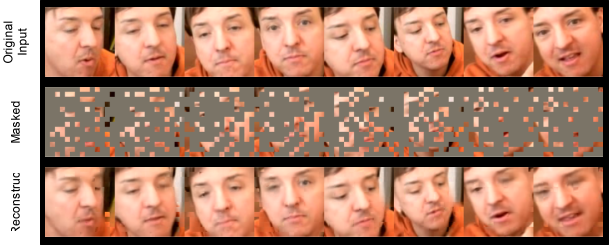}
      \caption{Reconstruction on ChaLearn First Impressions.}
      \label{fig:sfig2}
    \end{subfigure}
  \caption{Social-MAE visual zero-shot reconstruction on CREMA-D and ChaLearn First Impressions datasets. The first row shows the original input, the second row the visual equivalent to masked tokens, and the last row the reconstructed frames.}
  \label{fig:losses}
\end{figure*}

\subsection{Self-Supervised Pre-training}
\label{METH:SSL}
We adapted the pre-trained CAV-MAE model by training with self-supervision on the VoxCeleb2 dataset~\cite{chung2018voxceleb2}. VoxCeleb2 is an audiovisual dataset that contains over a million utterances from more than 6,000 speakers of 145 different nationalities. It provides a wide range of languages, accents, ethnicities and ages from real-world recordings. As self-supervised pre-training often requires vast amounts of data, we chose VoxCeleb2, as a suitable large and diverse audiovisual dataset with social content. 

The learning phase relies on the weighted combination of contrastive and reconstruction loss that provides complementary information. For an input sequence of N pairs of audio and video tokens {$a_i$, $v_i$}, the contrastive loss $\mathcal{L}_c$ is computed on modality averaged tokens {$c^a_i$, $c^v_i$} and aims to leverage relevant inter-modal information by following a LogSoftmax loss. 
The reconstruction loss $\mathcal{L}_r$ evaluates the model's ability to reconstruct masked tokens $x^{mask}_i$ from the tokens at the output of the decoder $\widehat{M}_i$ with an MSE loss. 
The final loss is the weighted sum of the contrastive and the reconstruction losses: $\mathcal{L} = \mathcal{L}_c \cdot \lambda_c + \mathcal{L}_r$.

\section{EXPERIMENTS AND RESULTS}

We pre-trained our Social-MAE during 25 epochs with a learning rate starting at $10^{-4}$ and decreasing at a decay rate of 0.5 every 5 epochs with a masking ratio p=75\%. For comparison, we also pre-trained CAV-MAE (as it uses 1 frame instead of 8) following the same settings. Both models were initialized on CAV-MAE$^{scale+}$ weights pre-trained on AudioSet-2M with self-supervision. We report visual zero-shot reconstruction in Fig.~\ref{fig:losses} using pre-trained Social-MAE on two downstream task datasets: CREMA-D~\cite{cao2014crema} and ChaLearn First Impressions (FI)~\cite{ponce-lopezChaLearnLAP20162016}. The model is able to provide a convincing output on previously unseen data. Most reconstruction errors, although not obvious at first sight, come from the most dynamic areas of the face, such as the eyes or lips. 



For downstream tasks, we remove the decoder from the architecture and replace it with a randomly initialized linear layer. We evaluate our pre-trained model by fine-tuning it on three different social and affective tasks: emotions recognition on CREMA-D~\cite{cao2014crema}, personality traits regression on ChaLearn FI~\cite{ponce-lopezChaLearnLAP20162016} and smiles and laughter detection on NDC-ME~\cite{heronDyadicConversationDataset2018}. For each task, we describe the dataset, the fine-tuning pipeline and the evaluation metrics to compare CAV-MAE and Social-MAE models against published baselines, following their experimental settings for consistency.

\subsection{Emotion Recognition}
\label{EMOREC}
\subsubsection{Experimental setup}
This task is evaluated on the Crowd-sourced Emotional Multimodal Actors Dataset (CREMA-D), containing 7,442 clips from 48 male and 43 female actors (20-74 years old). Each actor spoke 12 sentences using one of six emotions (anger, disgust, fear, happiness, sadness and neutral) ranging from 763 to 2204 utterances per emotion. Fine-tuning requires no masking on audio and visual tokens. We fine-tune pre-trained Social-MAE as well as our pre-trained version of CAV-MAE for 20 epochs using a mini-batch size of 8, learning rates at $10^{-4}$ and $10^{-5}$ for the encoders and the head respectively and we use the Cross-Entropy Loss.

\subsubsection{Baselines}
\paragraph{UAVM}
\cite{gong2022uavm} presented UAVM, a unified audiovisual framework for classification. The model uses pre-trained CNN-based feature extractors on log Mel filterbanks and multi-frame visual inputs that are fed to Transformer layers. 

\paragraph{AuxFormer}
\cite{goncalvesAuxFormerRobustApproach2022} proposed AuxFormer, a multimodal model that fuses audio and visual tokens through Transformer inputs. The model also processes separate modalities through auxiliary networks. The model loss is a weighted combination of the network losses. Audio inputs are low-level descriptors from OpenSmile~\cite{eybenOpensmileMunichVersatile2010} toolkit, and visual inputs are face clips processed by pre-trained VGG-face architecture~\cite{parkhiDeepFaceRecognition2015a}. 

\paragraph{VAVL}
\cite{goncalves2023versatile} proposed an audiovisual model, named Versatile AudioVisual Learning (VAVL), which relies on the Conformer architecture~\cite{gulati2020conformer}. Each modality input flows through a separate encoder followed by a shared-weight conformer. Audio inputs are high-dimensional features from Wav2vec2.0~\cite{hsu2021robust} and visual inputs are face clips processed into emotional feature representations. 

\begin{table}[htpb]
\caption{F1-score performance Comparison on CREMA-D. Mi and Ma refer to F1-score Micro and Macro. The best results are in \textbf{bold} face font. *~p-value $<$ 1E-5}
\label{tab:CREMAD}
\centering
\begin{tabular}{cp{.6cm}p{.6cm}p{.6cm}p{.6cm}p{.6cm}p{.6cm}}
                            & \multicolumn{2}{c}{\textbf{Audio}}            & \multicolumn{2}{c}{\textbf{Visual}}           & \multicolumn{2}{c}{\textbf{AV}} \\ \cline{2-7} 
 & \textbf{Mi} & \textbf{Ma}               & \textbf{Mi}               & \textbf{Ma} & \textbf{Mi}     & \textbf{Ma}      \\ \hline
AuxFormer~\cite{goncalvesAuxFormerRobustApproach2022}                   & 0.648          & 0.593 & 0.626 & 0.560          & 0.763              & 0.698               \\
UAVM~\cite{gong2022uavm}                        & 0.554          & 0.614 & 0.672 & 0.617          & 0.769              & 0.749               \\
VAVL~\cite{goncalves2023versatile} & \textbf{0.701}          & 0.628                & \textbf{0.787}          & 0.738       & 0.826              & 0.779               \\ \hline
CAV-MAE   & 0.694           & \textbf{0.694}                         & 0.630                         & 0.635           & 0.766             & 0.759               \\
Social-MAE                                     & 0.601*         & 0.607*                        & 0.749*                        & \textbf{0.755*}          & \textbf{0.837*}             & \textbf{0.842*}               \\ \hline 
\end{tabular}
\vspace{-5pt}
\end{table}

\subsubsection{Results and discussion}
Table~\ref{tab:CREMAD} reports the F1-score with micro and macro averaging techniques.  Social-MAE outperforms previously published methods for audiovisual classification. The micro F1 score shows the global accuracy, and the macro F1 score shows the unweighted average accuracy across each class, so the macro F1 score can be influenced by class imbalance. Since classes in CREMA-D range from 763 utterances (Sadness) to 2204 utterances (Neutral), we interpret the similarities between the macro and micro F1-scores reached by our pre-trained models as their ability to recognize emotions regardless of their prevalence.

Adapted CAV-MAE competes for best audio-only classification against VAVL model. 
Social-MAE rivals the best baseline for visual classification. The performance is impressive when you consider that the former processes 8 frames and the latter processes high-level features from all input frames. We also find it interesting that adapted CAV-MAE is able to outperform multi-frame baselines AuxFormer and UAVM on both unimodal and multimodal classification tasks, highlighting the efficiency of in-domain self-supervised pre-training.

\subsection{Personality Trait Prediction}
\label{PERSOTRAIT}
\subsubsection{Experimental setup}
We evaluate personliaty prediction with the First Impressions (FI) dataset, a collection of 10,000 \textit{in-the-wild} videos, in average 15s long. Videos are annotated with apparent personality traits known as \textit{big-5}~\cite{goldbergAlternativeDescriptionPersonality2001}: Openness, Conscientiousness, Extraversion, Agreeableness and Neuroticism. Fine-tuning requires no masking on audio and visual tokens. We fine-tuned both models presented in Sec.~\ref{METH:SSL} for 10 epochs using a mini-batch size of 8, an encoder learning rate of 1e-4 and the classification head learning rate of 1e-5. We use a Mean Absolute Error loss and our accuracy metric is $1 - \text{Mean Absolute Error}$.

We compare our fine-tuned CAV-MAE and Social-MAE to the best team of the challenge associated to the dataset: 
NJU-LAMDA~\cite{zhangDeepBimodalRegression2016}, a model pre-trained on VGG-face. The audio input is log Mel filterbank  and the visual input is the deep features from 100 frames. They train their model in 100 epochs for the audio stream and 3 epochs for the pre-trained visual stream, with a mini-batch of 128.

\begin{table}[htpb]
\caption{Model Accuracy on First Impressions Dataset. Best results are in \textbf{bold} face font. *~p-value $<$ 1E-5.
}
\label{tab:FI}
\centering
\begin{tabular}{cp{.6cm}p{.6cm}p{.6cm}p{.6cm}p{.6cm}p{.6cm}}
                            & \textbf{Ope.} & \textbf{Con.} & \textbf{Ext.} & \textbf{Agr.} & \textbf{Neu.} & \textbf{Avg.} \\ \hline
                            NJU-LAMDA~\cite{zhangDeepBimodalRegression2016}  & \textbf{0.912}    & \textbf{0.916}             & \textbf{0.913}        & \textbf{0.913}         & \textbf{0.910}       & \textbf{0.913}   \\ 
                            CAV-MAE    & 0.899         & 0.899             & 0.899        & 0.902         & 0.896            & 0.899        \\
                            Social-MAE & 0.908*    & 0.902*             & 0.895*        & 0.907*         & 0.905*       & 0.903*        \\ \hline
\end{tabular}
\vspace{-5pt}
\end{table}

\subsubsection{Results and discussion}
Table~\ref{tab:FI} shows the accuracy of each personality trait on ChaLearn First Impressions dataset as well as the mean accuracy. Social-MAE shows a performance of 90.32\% on average. While the accuracy is lower than the baseline, it remains impressive considering it was trained for only 10 epochs and with a smaller mini-batch size. 
We can also observe that processing multiple frames simultaneously (Social-MAE) demonstrates better regressions on four out of five traits compared to the single frame method (CAV-MAE).

\subsection{Smiles and Laughter Detection}
\label{LSN}
\subsubsection{Experimental setup}
The Naturalistic Dyadic Conversation on Moral Emotions (NDC-ME) dataset contains 8,352 clips of interactions in English of participants from different backgrounds. Each clip lasts 1.22 seconds, is cropped around the face, and is annotated with non-verbal expressions of smile, laughter, and neutral. We fine-tuned for 10 epochs, with no masking strategy, a mini-batch of 8, and learning rates of 1e-5 and 1e-4 for the backbone and classification head, respectively. Our training objective is the Cross-Entropy Loss. The baseline for smile and laughter detection is LSN-TCN~\cite{bohyNewPerspectiveSmiling2022a}, a CNN-based architecture that processes embedded representations of audio and video input separately and feeds them to two fully-connected joint layers. 

\begin{table}[htpb]
\caption{F1-score on NDC-ME. Best results are in \textbf{bold} face font. *~p-value $<$ 1E-5.}
\label{tab:NDCME}
\centering
\begin{tabular}{c|c|ccc}
                            & \textbf{Pre-training} & \textbf{Audio}          & \textbf{Visual}           & \textbf{Audiovisual} \\ \hline
LSN-TCN~\cite{bohyNewPerspectiveSmiling2022a}  & Supervised & 0.438            & 0.608           & 0.590            \\
CAV-MAE  & Self-Supervised & 0.471           & 0.629          & 0.766             \\
Social-MAE     & Self-Supervised & \textbf{0.546*}          & \textbf{0.728*} & \textbf{0.776*}             \\ \hline
\end{tabular}
\vspace{-5pt}
\end{table}

\subsubsection{Results and discussion}
Table~\ref{tab:NDCME} shows that both self-supervised methods reach higher F1-scores than the supervised baseline. Using multiple frames instead of one significantly improves the performance of the visual modality while slightly improving that of the multimodal classification.

\section{CONCLUSIONS}
In this paper, we presented Social-MAE, our pre-trained audiovisual Masked AutoEncoder on audiovisual social data. We modified existing CAV-MAE to accept multiple frames on a large human social behavior dataset. We evaluated our model on three relevant downstream tasks, demonstrating its effectiveness in achieving state-of-the-art results in audiovisual emotion recognition with a 0.837 F1 score and laughter detection with a 0.776 F1 score. With this work, we demonstrated the significance of in-domain adaptation of a large multimodal model trained through self-supervised pre-training. The proposed pre-trained encoder can be easily fine-tuned for other audiovisual social behavior understanding tasks, enabling more robust and performant models for perceiving human behaviors. 


\newpage

{\small
\bibliographystyle{ieee}
\bibliography{SocialMAE}
}

\addtolength{\textheight}{-3cm}   
                                  
\end{document}